\documentclass[10pt,twocolumn,letterpaper]{article}

\usepackage{iccv}

\definecolor{iccvblue}{rgb}{0.21,0.49,0.74}
\usepackage[pagebackref,breaklinks,colorlinks,allcolors=iccvblue]{hyperref}

\usepackage{amsmath,amsfonts,bm}

\def\eqref#1{equation~\ref{#1}}

\def\1{\bm{1}}

\DeclareMathAlphabet{\mathsfit}{\encodingdefault}{\sfdefault}{m}{sl}
\SetMathAlphabet{\mathsfit}{bold}{\encodingdefault}{\sfdefault}{bx}{n}

\usepackage{eccvabbrv}
\usepackage{hyperref}
\usepackage{url}
\usepackage{booktabs} 
\usepackage{graphicx}
\usepackage{booktabs}
\usepackage{multirow}
\usepackage{amsmath}
\usepackage{comment}

\usepackage{capt-of}  

\newcommand{\methodname}{CRAM (ours)}  
\newcommand{\vanillaname}{Upper Bound}  

\renewcommand{\paragraph}[1]{\vspace{2pt plus 1pt minus 1pt}\noindent{\bf #1}\;}

\newcommand{\ts}{\textsuperscript}

\begin{document}

\title{CRAM: Large-scale Video Continual Learning with Bootstrapped Compression}

\author{Shivani Mall \qquad Jo\~{a}o F. Henriques\\
Visual Geometry Group, University of Oxford\\
{\tt\small shivanim@robots.ox.ac.uk}
}

\maketitle

\begin{abstract}
Continual learning (CL) promises to allow neural networks to learn from continuous streams of inputs, instead of IID (independent and identically distributed) sampling, which requires random access to a full dataset. This would allow for much smaller storage requirements and self-sufficiency of deployed systems that cope with natural distribution shifts, similarly to biological learning.
We focus on video CL employing a rehearsal-based approach, which reinforces past samples from a memory buffer. We posit that part of the reason why practical video CL is challenging is the high memory requirements of video, further exacerbated by long-videos and continual streams, which are at odds with the common rehearsal-buffer size constraints. To address this, we propose to use compressed vision, i.e. store video codes (embeddings) instead of raw inputs, and train a video classifier by IID sampling from this rolling buffer. Training a video compressor online (so not depending on any pre-trained networks) means that it is also subject to catastrophic forgetting. We propose a scheme to deal with this forgetting by refreshing video codes, which requires careful decompression with a previous version of the network and recompression with a new one. We name our method Continually Refreshed Amodal Memory (CRAM). We expand current video CL benchmarks to large-scale settings, namely EpicKitchens-100 and Kinetics-700, storing thousands of relatively long videos in under 2 GB, and demonstrate empirically that our video CL method outperforms prior art with a significantly reduced memory footprint.

\end{abstract}

\section{Introduction}
\label{sec:intro}

Our world evolves endlessly over time. This temporal evolution creates a continuous shift in real-world data distributions. Crucially, resource-constrained autonomous agents must cope with these ongoing changes, akin to humans. Continual learning (CL) offers a practical solution to robustly acquire knowledge in non-stationary environments while amortizing the learning process over the agent’s lifespan \citep{thrun1995lifelong}. We focus on CL, utilizing long-video understanding as a proxy task for the real-world complexities encountered in actual deployment scenarios. Most existing CL research is restricted to static images or shorter video clips \cite{erace, pfnr, saha}, thus failing to model the natural shift in data distribution over extended time scales. In this work, we focus on naturally-collected long videos, which we believe is necessary to capture temporal progression and long distribution tails \cite{epickitchens}, challenges inherent to online learning \citep{catastrophic}. 

\begin{figure}[t]
    \centering \includegraphics[width=\linewidth]{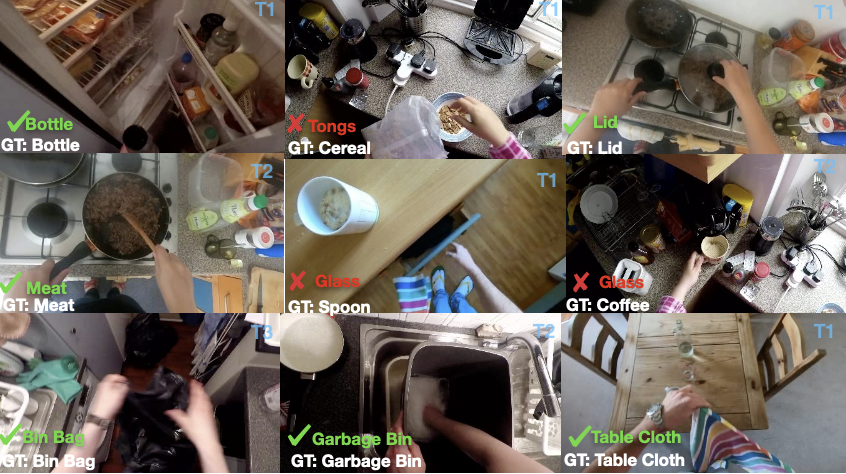}
    \caption{Visualization of the long-video continual learning setting. The model is exposed to a single long video over time, and must learn to predict new classes as they appear. The image shows 9 frames from one video, with predicted and ground truth classes. Note that predictions are not expected to be correct for new classes.}
    \label{fig:plot0}
\end{figure}

The extra temporal axis of video, compared to a static image, can capture rich information such as long-term activities. However, it also brings orders of magnitude more data, with the associated costs in processing and memory requirements \citep{htd}.
This challenge is made worse in the CL setting. Some of the most successful CL methods require random sampling from a rehearsal buffer of previous samples, with the finding that performance correlates with buffer size (sec. \ref{sec:related}). Large buffer sizes then quickly become prohibitively expensive if one is to consider long videos.

In this paper, we propose a memory-based video CL method that learns from naturally-collected long videos. Specifically, our method trains online a video compressor that continuously compresses and decompresses a neural-code rehearsal buffer, and an online classifier then uses the rehearsal buffer to learn video classification in the compressed space (sec. \ref{sec:method}). Unlike prior works, our rehearsal buffer stores compressed instead of raw video clips.
We mitigate forgetting in the compressor model with a ``buffer refresh'' scheme that only requires constant memory.

We draw inspiration from research into the mammalian brain and human dreaming, though like most works in CL we cannot claim biological plausibility \cite{ji2020automatic}. Specifically, hippocampal indexing theory states that the hippocampus stores compressed representations of neocortical activity patterns while awake \citep{mem_or, mem_cp}.
Temporal compression of events in episodic memory enables efficient storage and recall of past experiences \citep{mem1, mem2}, retaining information over long time periods \citep{mem3}.
Furthermore, an attractive hypothesis in neuroscience \citep{dreams} posits that dreams evolved to assist generalization and reduce forgetting. The hallucinatory and narrative nature of dreams potentially contribute to refining generative models, enhancing the brain's predictive processing capabilities (where predictions traverse top-down, while sensory input, bottom-up), and improving predictions about future states \citep{d1, d2, d3, d4}.
Our method's continuous compression and decompression of an episodic memory is inspired by a bottom-up/top-down enforcement of representation stability (though we make no claims of biological plausibility, given our use of back-propagation \citep{1, 2, 3}).


Our key contributions are as follows:
\begin{enumerate}
    \item A neural-code memory-based video continual learning framework that operates on long videos.
    \item A code refreshing scheme that minimizes representation drift in a buffer of codes that were initially created with different versions of the same compressor.
    \item An evaluation protocol for video CL in large-scale video datasets, namely Epic-Kitchens-100 and  Kinetics-700.
    \item Empirical evaluations in both datasets, in 2 CL settings (with pre-training, and incrementally from scratch). Experiments show the proposed method's performance for videos over $10\times$ longer than previously demonstrated.
\end{enumerate}

\section{Related Work} \label{sec:related} \paragraph{Continual Learning with Images and Videos.} 
Most current CL systems show promising results in the image domain, which primarily involves artificially-constructed sequences of images and transfer of declarative knowledge of entities and concepts \citep{pb, all_cl, paz, ji2020automatic}.
Some works in CL operate on videos; however, they are limited to processing only videos with a few seconds or minutes, or do not propose scalable approaches to tackle the high memory and computational requirements. FrameMaker \cite{framemaker} addressed memory efficiency by introducing frame condensing, where a single condensed frame per video is stored, along with instance-specific prompts. However, it uses pretrained ImageNet initializations, which does not support incremental learning from scratch. Additionally, there is significant overlap between ImageNet classes and Kinetics-700 classes, which
makes it difficult to create a train-test split for CL classes.
OAK (Wanderlust) \citep{oak} proposed a benchmark with long ego-centric videos, but limited to a
narrow task domain: coarse-grained object detection with sparse annotations.
This benchmark was also used in Efficient-CLS \citep{ecls} which proposed a slow-fast CL method with an episodic memory similar to prior work \citep{reb, paz, chau}.
Based on Complementary Learning Systems \citep{cls}, Efficient-CLS \citep{ecls} is complementary to other CL methods, augmenting them with a pair of slow and fast learners, and using the former to generate pseudo-labels for the later.
\Citet{ecls} also shows performance on EgoObjects \citep{egoobj}, a fine-grained ego-centric dataset with seconds-long clips. This contrasts with our experiments using Epic-Kitchens-100 \citep{epickitchens}, with minutes to hours-long videos.
CLAD \citep{clad}, a CL benchmark for autonomous driving, repurposed an image dataset to form a temporal stream. It proposed a single (days-long) ``video'' (time lapse sequence of images), with domain shifts at different frequencies (\eg time, location, different objects, viewpoint). While a single very long video is a reasonable axis to expand video CL evaluation, we extend it to thousands of videos, each with minutes or hours \citep{epickitchens}.

\paragraph{Memory-Based Continual Learning.} Memory-based algorithms have strong performance in CL \citep{saha, gd22, chau}. During training, a memory buffer stores samples from the past and rehearses them while training new tasks, mitigating catastrophic forgetting. \citet{mem_cp} proposed a compression-based CL method over static images and natural language, but did not consider long videos. Other works primarily focus on different memory budgets, balancing or rehearsal \citep{gd22}, which is orthogonal to compression.

\paragraph{Rehearsal-Free Continual Learning} DPAT \cite{dpat}, ST-Prompt \cite{st-prompt} and L2P \cite{l2p} achieve CL without rehearsal by using \emph{pretrained} large-scale vision/vision-language models and prompts to encode sequential information, thus sidestepping the need for a memory buffer.
However, such large-scale models
can consume up to several hundred gigabytes, limiting their practical usage in edge computing.

\paragraph{Video Compression.} Training video representations is, in a sense, more challenging than image representations, due to the enormous size of raw video streams and the high temporal redundancy.
Neural video compression was shown to reduce superfluous information by up to two orders of magnitude \citep{wu, oliviacv}.
An important benefit is that compressed video representations can act as generic features extractors, reducing capacity requirements of a downstream model.
Efficient video learning remains a hard task, and several works proposed performance improvements such as token dropout, frame sampling and key information detection \citep{yan, htd, zhi}.
Neural compression presents an elegant solution for these challenges \citep{wu}.


\paragraph{Robot Lifelong Learning.} A research direction in robotics explores CL methodologies utilizing videos and feedback mechanisms \citep{thrun1995lifelong, stone}. In this area, robots are tasked with acquiring and refining their skills and knowledge over time.
Robot lifelong learning typically focuses on active learning and the effect of an agent's actions in the environment, which typically adds actions as an output modality and reinforcement or imitation learning as a goal \cite{lib}.

\section{Background}\label{sec:background}

\subsection{Compressed Vision}
Our method builds on compressed vision, proposed by Wiles \etal \citep{oliviacv}. The main concept is to train any classifier on small codes (embeddings) obtained from video frames, instead of the frames directly.
By using a frozen compressor network to obtain the codes, and performing data augmentation (to avoid overfitting) directly in the code latent space instead of the input space, they can store extremely long videos in memory compared to traditional approaches.
Their pipeline consists of three training phases.
(1) They train a \emph{neural compressor} $c = (\phi, \psi)$, where $\phi$ and $\psi$ denotes the encoder and decoder respectively, using a VQ-VAE \citep{vqv17}.
$c$ takes videos $X$ as input and produces neural codes $x \in \mathbb{R}^{s\times h\times w}$.
After this phase, the neural codes are stored in a buffer, $c$ is frozen and the original videos are no longer needed.
(2) They train an augmenter network $a$, that takes as input $x$ and predicts codes $\hat{x_i}$ that correspond to randomly-transformed video frames.
(3) Lastly, they train a video task classifier that takes as input $\hat{x}$ to solve a given downstream task, and prevent over-fitting by using $a$ to perform data augmentation directly in the space of the codes.
Wiles \etal \citep{oliviacv} show strong performance results (under 5\% accuracy drop) at high compression rates ($256\times$ and $475\times$).

\subsection{Incremental Learning}
A common scenario in CL \citep{chau, reb, paz} is incremental learning -- training a network by presenting it with a sequence of $n$ tasks consisting of disjoint data distributions, sequentially, as $T = \left\{t_{i} \right\}_{i}^{n}$. This models a shifting data distribution as a sequence of distributions.
Concretely, a learning model observes a \emph{continuum of data}, which is a concatenation of $m$ samples from each task, for a total of $nm$ samples:
\begin{alignat}{1}
\label{eq:continuum_data}
& D = \left\{x_{j,i}, y_{j, i} \right\}_{j,i}^{m,n} \\
& x_{j,i} \stackrel{iid}{\sim} X_{t_i}, \quad y_{j,i} \stackrel{iid}{\sim} Y_{t_i}
\end{alignat}
$X_{t_i}$ is a distribution over images for task $t_i$, and $Y_{t_i}$ is a distribution over its target vectors (for example, action classes). For simplicity, we assume samples are IID within a task.

The main advantage of this setting is that it represents the most stringent test of continual learning, by training from scratch. It also more closely mimics biological learning, i.e. an agent learning solely from sequential experience.

\subsection{Pre-training and Incremental Learning}
A variation of incremental learning is to allow an initial pre-training phase \citep{podnet}, where the network is trained on a large subset of the classes (e.g. half of them) IID, and then is incrementally adapted as before.
This more closely follows a common usage of ML models, where there can be some relevant dataset for pre-training before deploying a system.

\begin{figure*}
    \centering
    \includegraphics[width=0.80\linewidth]{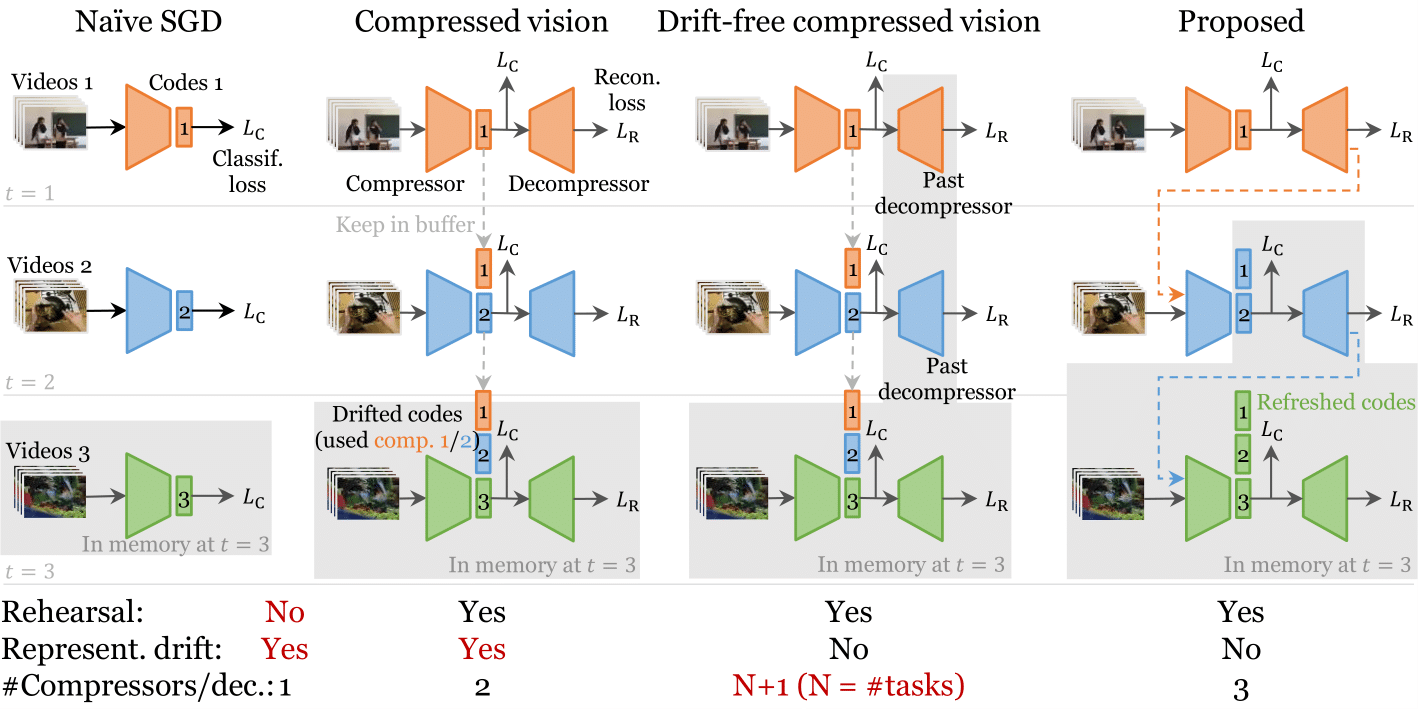}
    \vspace{-5pt}
    \caption{Overview of the differences between our proposed scheme and alternative compressed buffer strategies. Using a compressed buffer for rehearsal (column 2) risks representation drift, since codes were created with a different version of the trained encoder (represented as 3 different colors). Decoding without drift requires snapshots of the decoder over time (column 3), but the memory growth is unbounded. Our proposal (column 4) refreshes codes to keep them from drifting, while only requiring a single snapshot of the last decoder.}
    \label{fig:differences}
\end{figure*}

\section{Method} \label{sec:method}


As in sec. \ref{sec:background}, we aim to train a deep network by presenting it with a sequence of $n$ tasks of disjoint data distributions, i.e. eq. \ref{eq:continuum_data}.
The main difference is that each $x_{j,i}$ is a video clip, and each $y_{j,i}$ is now a video class (e.g. an action label).

\subsection{The Ideal Case: IID Sampling}
We will first present the ideal case, where a learner has access to all available samples, sampled IID. This avoids catastrophic forgetting and allows us to introduce the concepts in a simplified form.
We aim to train a feature extractor or compressor $c = (\phi, \psi)$, composed of an encoder $\phi$ and decoder $\psi$, as well as a classifier $q$ which takes the features from the encoder.
The  objective of the compressor, trained on the full dataset from eq. \ref{eq:continuum_data}, is defined as:
\begin{equation}\label{eq:iid-compression-objective}
    \psi^\ast,\phi^\ast = \arg\min_{\psi,\phi} \left(\underset{t_i \sim T}{\mathbb{E}} \; \underset{x_{j} \sim X_{t_i}}{\mathbb{E}}\left(||\psi(\phi(x_{j})) - x_{j}||^2\right)\right).
\end{equation}
The classifier is trained with a cross-entropy loss $L$ for classification (or another loss for a different downstream task):
\begin{equation}\label{eq:iid-classification-objective}
    q^\ast = \arg\min_q \underset{t_i \sim T}{\mathbb{E}} \; \left( \underset{(x_{j},y_{j}) \sim (X_{t_i},Y_{t_i})}{\mathbb{E}} L(q(\phi(x_j)), y_j) \right)
\end{equation}
This is, of course, an idealized situation where it is possible to have random access to any sample. Next we will turn to the CL scenario where we are given only a single task $t_i$ at a time, and cannot directly access past samples.
A naive CL baseline is illustrated in the 1\ts{st} column of fig. \ref{fig:differences}.


\subsection{Incremental Learning}
In this setting, the compressor is trained continually with new classes. Unlike \citep{oliviacv}, as time progresses, the compressor observes new data samples unseen during past tasks, and as such is also subject to catastrophic forgetting.
This naive baseline is illustrated in the 2\ts{nd} column of fig. \ref{fig:differences}.

\subsubsection{Rehearsal Buffer and Temporal Model Evolution}
Because we train a model sequentially over the tasks, and it will be different for each task, we need to consider a \emph{sequence of models} $(c_1, q_1), \ldots, (c_n, q_n)$, one per task $t_i$.

In order to allow training on past samples, so that the loss value on them is maintained, some form of memory (explicit or implicit) is also required.
In this work we maintain a buffer denoted as $B_{i-1}$.
At time $t_i$ it is defined as
\begin{equation}
\label{eq:buffer}
B_{i-1} = \left\{e_{j,k} \right\}_{j,k}^{m,i-1},\quad e_{j,k} = \phi_{t_{i-1}}(x_{j,k})
\end{equation}
where $k$ iterates over previous tasks (1 to $i-1$), $j$ iterates over samples per task (1 to $m$), and $e_{j,i}$ denotes the compressed video clip.
The \emph{neural codes} buffer $B_{i-1}$ contains previously observed video codes, necessary to maintain concepts from prior tasks.
During task $t_i$, when training $c_i$ and $q_i$, we only have access to the last state of the buffer $B_{i-1}$ and video examples from the current task, $X_{t_i}$.

A major issue arises if we naively merge rehearsal-based CL with a buffer of compressed codes:
since the compressor $c$ changes with time, the codes will get ``stale'' over time (as they were encoded with a previous version of the compressor).
This is illustrated in the 2\ts{nd} column of fig. \ref{fig:differences} (notice the codes with different colors, corresponding to different compressor versions).
Keeping all decompressor versions would address this, but at a large memory cost (3\ts{rd} column of fig. \ref{fig:differences}).
The next section will present a solution.

\subsubsection{Incremental Learning Formulation}\label{sec:incremental-formulation}
%
Let us consider the first task. Adapting eq. \ref{eq:iid-compression-objective} to focus on it:
\begin{equation}\label{eq:compress-time1}
    \psi^\ast_1, \phi^\ast_1 = \arg\min_{\psi_1, \phi_1} \; \underset{x_{j} \sim X_{t_1}}{\mathbb{E}}\left(||\psi_1(\phi_1(x_{j})) - x_{j}||^2\right),
\end{equation}
and an identical adaptation for the classifier from eq. \ref{eq:iid-classification-objective}.
Similarly, for the second task, we have the loss equation:
\vspace{-1em}
\begin{alignat}{1}
    \psi^\ast_2, \phi^\ast_2 =& \arg \min_{\psi_2, \phi_2} \left(\underset{x_{j} \sim X_{t_2}}{\mathbb{E}}\left(||\psi_2(\phi_2(x_{j})) - x_{j}||^2\right)\right. \nonumber \\
    & + \left. \underset{e_{j} \sim B_{1}}{\mathbb{E}}\left(||\psi_2(\phi_2(s_{j})) - s_{j}||^2\right)\right) \label{eq:compress-time2} \\
    & \textrm{where } s_{j} = \psi_1(e_{j})\label{eq:refreshed-code}
\end{alignat}
where the first expectation is over the current batch, and the second expectation is over codes stored in the buffer, which are decoded by $\phi_1$. It is important to decompress the buffer using the \emph{decoder parameters from the previous task} $\psi_1$, not the one currently being trained $\psi_2$, in order to be consistent with the encoder they were compressed with,  $\phi_1$.

Note that we do not retain the decoders from all the previous CL tasks (3\ts{rd} column of fig. \ref{fig:differences}), but rather retain a \emph{single} one from the immediately-previous task. This allows reconstructing all the codes stored in the buffer, as long as we are careful to ``refresh'' stored codes every time they are loaded (by compressing them with the new compressor). This process is illustrated in the 4\ts{th} column of fig. \ref{fig:differences}.

As for the classification objective (eq. \ref{eq:iid-classification-objective}), it also considers codes from the buffer and samples from the current task:
\begin{alignat}{1}\label{eq:classify-time2}
    q^\ast_2 = \arg\min_q & \left( \underset{(x_{j},y_{j}) \sim (X_{t_2},Y_{t_2})}{\mathbb{E}} L(q(\phi_2(x_j)), y_j) \right. \nonumber \\
    & \quad + \left. \underset{(e_{j}, y_j) \sim B_{1}}{\mathbb{E}} L(q(\phi_2(s_j)), y_j) \right),
\end{alignat}
where we reuse eq. \ref{eq:refreshed-code}, and slightly abuse notation to retrieve the classification label $y_j$ associated with the buffer's code $e_j$. Eq. \ref{eq:classify-time2} is applied to each CL task after refreshing codes. Note the latest encoder refreshes \emph{all} codes in the buffer, and thus we can use the \emph{same} encoder for all tasks in eq. \ref{eq:classify-time2}.

\begin{figure}[t]
    \centering
\includegraphics[width=\columnwidth]{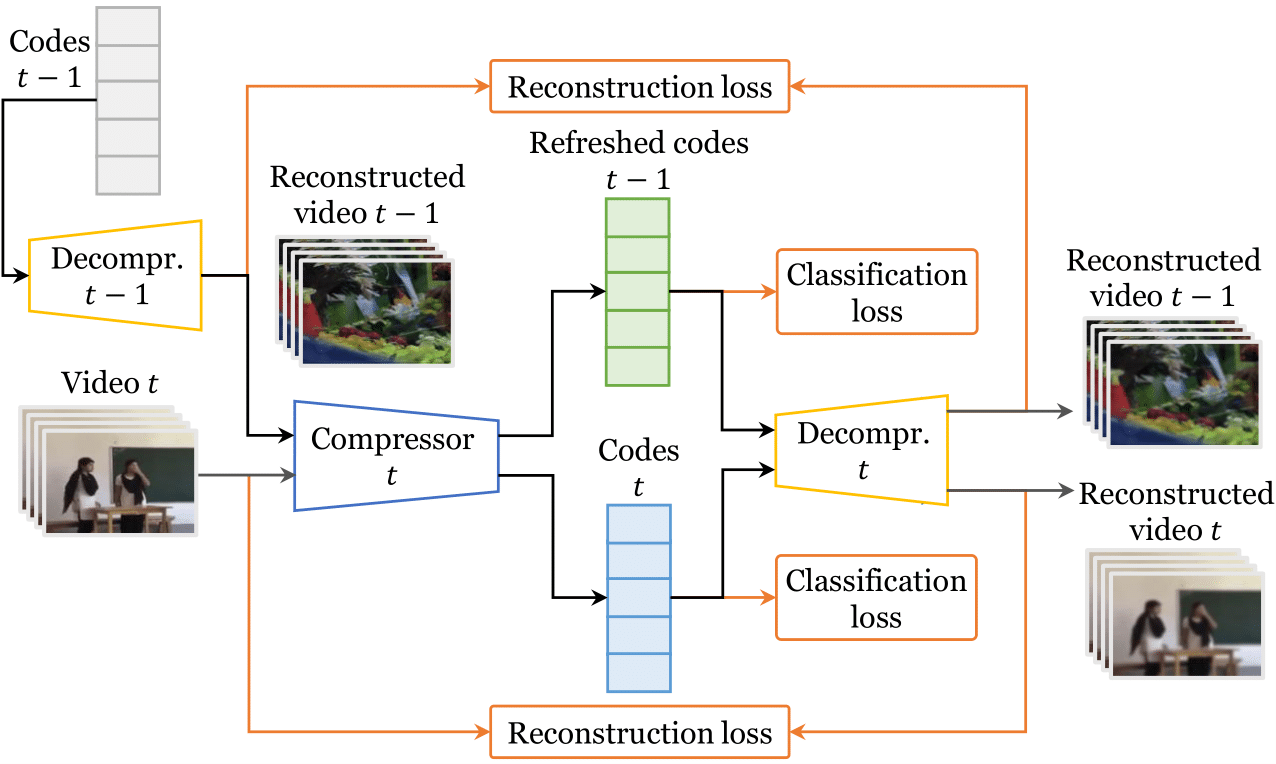}
    \caption{Overview of the proposed compressed continual learning pipeline. Our method trains a video compressor as an autoencoder, together with a classifier, while storing short compressed codes describing the videos in a buffer for rehearsal of past samples. Our method continually refreshes codes from past tasks $t-1$ so that they work with the compressor for the current task $t$, ensuring the stability of the representations over time.}
    \label{fig:pipeline}
\end{figure}

We can apply eq. \ref{eq:compress-time1}-\ref{eq:compress-time2} recursively to any task $t_k$ by using the buffer and compressors from the respective tasks, and thus extend it by induction.
Fig. \ref{fig:pipeline} shows an overview.
We name our method Continually Refreshed Amodal Memory (CRAM), since the memory is not in the input modality.

\begin{table*}[t]
\centering\footnotesize
\begin{tabular*}{\textwidth}{@{\extracolsep{\fill}}lccccccc}
\toprule
 &  & \multicolumn{3}{c}{Kinetics-700 (K-700)} & \multicolumn{3}{c}{EpicKitchens-100 (EK-100)}\\
 \cmidrule{3-5} \cmidrule{6-8}
Setting & Method & Train. $\uparrow$ & Eval. $\uparrow$ & AvgF $\downarrow$ & Train. $\uparrow$ & Eval. $\uparrow$ & AvgF $\downarrow$\\
\midrule 
\multirow{2}{*}[-2.5ex]{Pretraining} & \color{gray}\vanillaname{} & \color{gray}57.10\tiny{$\pm$0.7} & \color{gray}48.20\tiny{$\pm$0.8} & -- & \color{gray}42.10\tiny{$\pm$0.75} & \color{gray}35.90\tiny{$\pm$0.71} & --\\
\cmidrule{2-8} 
 & \methodname{} & \textbf{56.25\tiny{$\pm$0.5}} & \textbf{46.50\tiny{$\pm$0.41}} & \textbf{5.50\tiny{$\pm$0.35}} & \textbf{40.10\tiny{$\pm$0.37}} & \textbf{33.20\tiny{$\pm$0.38}} & \textbf{9.70\tiny{$\pm$0.21}} \\
\cmidrule{2-8} 
 & REMIND \cite{mem_cp} & 45.8\tiny{$\pm$0.57} & 38.20\tiny{$\pm$0.5} & 43.10\tiny{$\pm$0.5} & 32.3\tiny{$\pm$0.67} & 26.50\tiny{$\pm$0.56} & 51.0\tiny{$\pm$0.4} \\
\midrule 
\multirow{5}{*}[-3ex]{Incremental} & \color{gray}\vanillaname{} & \color{gray}48.20\tiny{$\pm$0.7} & \color{gray}44.10\tiny{$\pm$0.65} & -- & \color{gray}36.20\tiny{$\pm$0.55} & \color{gray}32.0\tiny{$\pm$0.47} & -- \\
\cmidrule{2-8} 
 & \methodname{} & \textbf{44.60\tiny{$\pm$0.25}} & \textbf{38.80\tiny{$\pm$0.2}} & \textbf{15.20\tiny{$\pm$0.1}} & \textbf{32.60\tiny{$\pm$0.27}} & \textbf{28.10\tiny{$\pm$0.3}} & \textbf{21.60\tiny{$\pm$0.15}} \\
\cmidrule{2-8} 
 & SMILE \cite{lama} & 40.56\tiny{$\pm$0.55} & 29.20\tiny{$\pm$0.47} & 62.50\tiny{$\pm$0.40} & 28.71\tiny{$\pm$0.68} & 19.20\tiny{$\pm$0.63} & 67.8\tiny{$\pm$0.60} \\
\cmidrule{2-8}
 & vCLIMB \cite{vclimb} & 39.12\tiny{$\pm$0.35} & 28.65\tiny{$\pm$0.3} & 65.10\tiny{$\pm$0.4} & 27.11\tiny{$\pm$0.8} & 18.5\tiny{$\pm$0.66} & 66.5\tiny{$\pm$0.5} \\
\cmidrule{2-8} 
 & GDumb \cite{gd22} & 37.61\tiny{$\pm$0.15} & 18.70\tiny{$\pm$0.25} & 52.40\tiny{$\pm$0.2} & 25.30\tiny{$\pm$0.35} & 15.60\tiny{$\pm$0.3} & 60.10\tiny{$\pm$0.27} \\
\bottomrule
\end{tabular*}
\caption{Comparison of our method and baselines (average training (Train) and evaluation accuracy (Eval), and average forgetting (AvgF)), on K-700 and EK-100, with pre-training and incremental settings (as described in Sec 5.2 and 5.3). 
We set 1404 Mb (in K-700) and 1464 Mb (in EK-100) as the maximum memory budget for our method and baseline experiments above (as described in Sec 5.4). Upper Bound refers to the upper bound baseline which has unbounded memory budget (described in Sec 5.4).}
\label{table:Tab1}
\end{table*}

\subsection{Continual Learning with Pre-training}
Another natural setting as illustrated in \citep{podnet} is to consider networks that undergo pre-training with IID samples prior to incremental learning.
In this setting, there are two phases. In the first phase, the model is pre-trained with half of the dataset's classes and in the second phase, the model is incrementally trained with rest of the classes.

Following the protocol of \citet{podnet}, in the first phase we pre-train the compressor and classifier with half of the dataset's classes, and in the second phase, we incrementally train the classifier as in sec. \ref{sec:incremental-formulation}. 

Note that an important distinction from the previous setting described in \ref{sec:incremental-formulation} is that after phase 1 finishes, we can freeze the compressor -- assuming that the pre-training is sufficient to learn relevant features -- and as a result, during phase 2, we do not decompress our buffered codes. This avoids representation drift of the codes and simplifies the method, by not needing to back-propagate through codes.

It is interesting to contrast this pre-training setting to the incremental learning only setting (sec. \ref{sec:incremental-formulation}).
Continuously decompressing and adapting the codes incurs a computational cost and risks representational drift.
Under a bounded memory budget, the compressor may be under-trained and fail to produce robust codes. The pre-training setting circumvents these issues, while still enjoying the benefits of incremental learning of downstream tasks.

\section{Experiments} \label{sec:experiments}

To demonstrate our method empirically, we evaluate on memory-based and video CL baselines. We use Kinetics-700 (K-700) \citep{will} and Epic-Kitchens-100 (EK-100) \citep{epickitchens}, for action and noun classification, respectively.

EK-100 and K-700 have approximately 16 and 14 million frames respectively for training, and 4 and 3 million frames respectively for evaluation. The longest video length in EK-100 is 1.5 hours, and 15 seconds in K-700. The average video length in EK-100 is approximately 20 minutes, and 10 seconds in K-700. We organize the data into a class-incremental setting, following the strategy introduced in the vCLIMB \cite{vclimb} benchmark guidelines.

\subsection{Implementation details}
We use the same compressor architecture as Wiles \etal \citep{oliviacv}, which is based on a ResNet, and refer the reader to their work for a complete review. Compressor training differs in the two settings as described below. In both the settings, we maintain a queue for the rehearsal buffer to store the video codes. For the downstream video task classifier, we use S3D \citep{xie} for K-700, and short-term S3D for EK-100, which takes the compressed codes as inputs. We follow the specifications from  Wiles \etal \citep{oliviacv} to adapt the network's kernel size and stride at every layer. We experimented with different architectures for the classification task, in order to find the optimal settings. We use compression rate $256\times$ unless stated otherwise. We apply random horizontal flipping and cropping of size $224\times 224$, from frames resized to have a short side $\in$ [256, 340].

Each video is split into clips of dimensions $224\times 224\times 3\times 32$ (32 RGB frames), and then compressed into a code that corresponds roughly to a size 0.0013 Mb. This is the size of each clip that we encode into a code, and each long video is composed of many such clips / codes.
Every block of frames within the video is compressed independently, instead of the entire video with one code. As each code encodes a few frames at a time, the number of codes varies depending on the video length. A temporally-long video has a higher number of neural codes for it in comparison to a short video. For Kinetics-700, a video (with 250 frames on an average) has approximately 8 codes associated with it whereas in Epic Kitchen-100, a video (with 27K frames on an average) has approximately 850 codes associated with it.

%

\subsection{Setting: Continual Learning with Pre-training}

\paragraph{Dataset.}
Following the experimental protocol in \citep{podnet}, we split K-700 into 2 parts. The first split consists of Kinetics-400 (K-400), and the second split contains the remaining 300 classes of K-700. Similarly, we split EK-100 into 2 parts, the first with 17 participants and the second with 16 participants. Classes are sampled IID in the first dataset split respectively. For K-700, the second split has 10 tasks with 30 non-overlapping classes per task. For EK-100, the second split has 17 tasks with 1 participant per task. Videos are sampled IID within every task. Further details on video samples storage in A.1 (app.).

\paragraph{Training.}  As described in Section 3, this setting has two phases. In the first phase, we follow IID training. We train the compressor $c$ and classifier for 300 epochs with a batch size of 32, and use the Adam optimizer with learning rate of $0.01$ and weight decay of $10^{-5}$. $c$ is frozen at the end of pre-training. We store the compressed codes into the queue for all the classes in this phase, and then train the classifier with these stored codes. We start the second phase with the pre-trained classifier from the first phase and train it incrementally over 10 tasks for K-700 and 17 tasks for EK-100. We pass the transformed video inputs through the frozen compressor, store the resulting codes into the queue and use them as inputs for the classifier. We receive new class samples at every task, and assume IID sampling over those. We train the classifier for 2 epochs with the compressed codes corresponding to new samples and those stored in the rehearsal buffer. Note that the buffer also includes the codes from pre-training classes, plus from all tasks seen so far. We also perform ablations varying the number of epochs per task and class splits. The incremental training over the classifier completes once all the tasks are processed.

\subsection{Setting: Incremental Learning from Scratch}

\paragraph{Dataset.}
For K-700, we have 35 tasks with 20 non-overlapping classes per task. For EK-100, we have 33 tasks with video samples from 1 participant per task. Videos are sampled IID within every task.

\paragraph{Training.} We train the compressor and classifier incrementally. Within each task, we perform IID sampling to first train the compressor for 1 epoch and then the classifier for 30 epochs, unless stated otherwise. To train the compressor, we use a batch size of 16, and Adam optimizer with learning rate of $0.01$ and weight decay of $10^{-5}$. At every task, during compressor training, we decompress compressed codes from the buffer (unless empty) using the latest compressor, and obtain the corresponding RGB values. We then re-train the compressor jointly with the decompressed codes and new samples. Lastly, we store the freshly compressed codes into the buffer, and freeze the compressor for that task. We use the stored codes from the current and past tasks as inputs to the task classifier. We store the resulting codes for every video clip into the queue, and freeze the compressor for that task. 
This training process is repeated for the total number of tasks.

\subsection{Baseline details} Our method lies at the intersection of memory and video CL, so we compare with GDumb \citep{gd22} and REMIND \citep{mem_cp}, (both are image-based memory CL methods), and SMILE \cite{lama} and vCLIMB \cite{vclimb} (both are video CL methods). We propose an extension of the image-based CL evaluations to video datasets. We also design an upper bound baseline using an unbounded RGB memory budget. To compare with REMIND \citep{mem_cp}, we use our pre-training setup, as it relies on a pre-trained architecture. For the other methods, we use our incremental learning setup. Further details on the baseline implementations are in B.2 (app.).

\subsubsection{Memory Budget}
In our baseline comparisons for K-700 and EK-100, we set 1404 Mb and 1464 Mb respectively for the maximum memory budget 
of all methods, in order to ensure a fair comparison.
These values were chosen as the maximum memory that the proposed method requires, and they are well within the capacity of modern hardware.

During the incremental learning phase, at every task, we split the storage space equally for each past task up to the buffer limit. Denote $K$ as the total number of video samples that can be stored under the assigned memory budget. Then the total number of samples from each past task at the $n^{th}$ task in the incremental learning setting is given by $K\frac{1}{n-1}$. The total number of samples from each past task at the $n^{th}$ task in the pre-training setting, where we add one task for the pre-training phase, is $\frac{K}{n}$.

\begin{figure}
    \centering
    \includegraphics[width=0.8\linewidth]{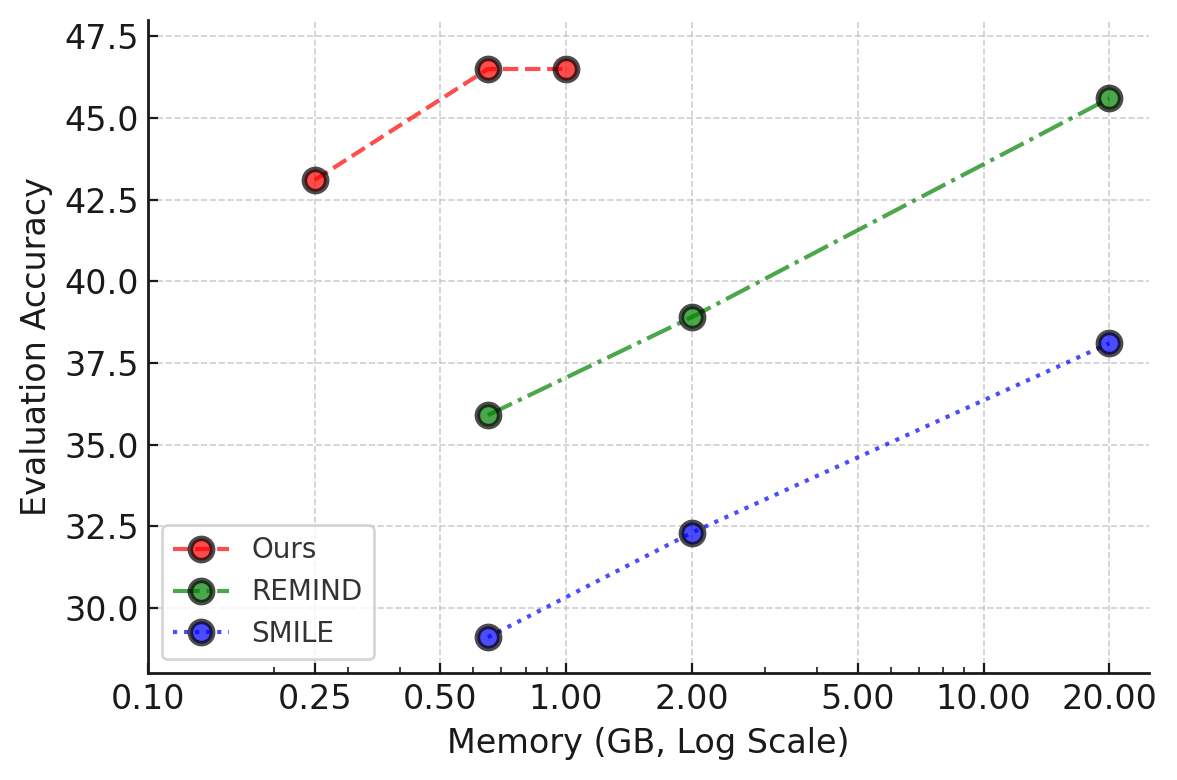}
    \caption{Average evaluation accuracy for different methods, with varying memory budgets on Kinetics-700.}
    \label{fig:plot}
\end{figure}

\begin{table}
\centering\scriptsize\setlength\tabcolsep{4pt}
\begin{tabular}{cccc}
\toprule 
Dataset & Memory (MB) & \methodname{} & RGB buffer (Upper Bound)\tabularnewline
\midrule 
 & Buffer & 654 & $1503\times10^{3}$ \tabularnewline
\cmidrule{2-4}
Kinetics-700 & Models & 750 & 250 \tabularnewline
\cmidrule{2-4} 
 & Total & \bf 1404 & $~1503.2\times10^{3}$\tabularnewline
\midrule 
 & Buffer & 714 & $1640\times10^{3}$ \tabularnewline
\cmidrule{2-4} 
EpicKitchens-100 & Models & 750 & 250 \tabularnewline
\cmidrule{2-4} 
 & Total & \bf 1464 & $1640.2\times10^{3}$\tabularnewline
\bottomrule
\end{tabular}
\captionof{table}{Memory footprint of our method with a compression ratio of 256$\times$ versus a traditional buffer of RGB images.}
\label{table:Tab2}
\end{table}

\begin{table}
\centering\scriptsize \setlength{\tabcolsep}{10pt}
\begin{tabular}{lcc}
\toprule
\textbf{Method} & \textbf{Kinetics-700} & \textbf{Epic-Kitchens-100} \\
\midrule
\bf{CRAM}  & \bf 1.4 × $\mathbf{10^7}$ & \bf 1.6 × $\mathbf{10^7}$  \\
\midrule 
REMIND & 7.5 × $10^5$ & 8.5 × $10^5$ \\
\midrule 
RGB buffer  & $\sim$3.0 × $10^3$ & $\sim$7.0 × $10^3$ \\
\bottomrule
\end{tabular}
\caption{
Number of frames stored with a constant memory budget. RGB buffer baselines include GDumb, SMILE and vCLIMB.
}
\label{tab:method_mem_perf}
\end{table}

\label{sec:results}
\subsection{Results}

In table \ref{table:Tab1}, we report the average forgetting (AvgF, defined in B.1, app.) and training / evaluation average accuracy (average on all the tasks measured at the conclusion of the task sequence) \citep{vincl}  for our method and baselines. We find that our method outperforms the baselines and achieves average accuracy comparable to the upper bound baseline in both our proposed settings, as seen in table \ref{table:Tab1}. We observe that our pre-trained compressor captures class-agnostic semantics effectively. For samples unseen during the pre-training phase, it outputs robust compressed codes without further training, thus enabling the online classifier to achieve strong performance. In the incremental learning only setting, at every successive task, since our method decompresses and rehearses past codes, it learns to jointly represent the features for both old and new tasks. This allows it to output robust codes for downstream video application. Due to the highly efficient memory, it enjoys full rehearsal of samples from all past tasks, thus our classifier can efficiently represent all classes, and achieve strong performance. We show some examples of our method's predictions, learned over time in fig. \ref{fig:plot0} (more examples in fig. 1 of the app.).

We observe that our compression strategy is well-optimized such that, even for videos (10x longer than previously demonstrated) with high memory footprint, we only need a small amount of memory ($<$ 2 GB). In table 1 (app.), we present a qualitative comparison for video datasets used in ours and previous works. Interestingly, unlike prior work, we do not need to apply \emph{any} frame selection or sampling strategy, even for very large videos, as our method's efficiency enables storing every video frame. Further, by extending evaluation to complex long videos (such as in Epic-Kitchens-100 \cite{epickitchens}), our work covers fine-grained tasks with hand-object manipulation posing full and partial occlusions, multiple viewpoints and distribution shifts not tackled by earlier works (table 1, app.).

\subsubsection{Discussion}
We propose a method that circumvents the need for frame selection, so metrics like memory frame capacity \cite{vclimb} used to evaluate frame selection strategies in video CL are not appropriate. Instead, average forgetting (AvF, tab. \ref{table:Tab1}) and total memory are better measures of efficiency, as in fig. \ref{fig:plot}.

Our work extends the evaluation to long videos settings, a setting in which prior work showed that frame selection or condensing is very detrimental to performance \cite{adv1, adv2}. This is due to loss in temporal continuity, crucial for fine-grained and long-term context preservation \cite{adv1, adv2}. In contrast, \citet{lama, framemaker, vclimb} focus on short-video settings, where adverse effects from frame selection are negligible due to simpler temporal dynamics.

\subsection{Baseline Results}
\paragraph{Memory-Based CL Baselines.}
We first compare with GDumb \citep{gd22}, a simple baseline that puts other works into context. Unlike prior art, we extend the evaluation to long videos, so GDumb serves as a sanity check to ensure we can genuinely outperform naive CL strategies. Due to insufficient samples for rehearsal, GDumb suffers from forgetting with a far lower accuracy than our method (table \ref{table:Tab1}).

Related to our method, REMIND \cite{mem_cp} focuses on compressing the raw RGB input using quantization and storing the resulting compressed codes instead, in the replay buffer. It also follows the benchmark setup described in sec. \ref{sec:method}. Our method achieves a 20x higher compression rate, and maintains a wider temporal history, outperforming REMIND on both datasets. Furthermore, it does not refresh representations but only the final layer features, which may explain the lower performance on demanding downstream applications.

\paragraph{Video CL Baselines.}From table \ref{table:Tab1}, we observe that SMILE \citep{lama} and vCLIMB \cite{vclimb} show significant performance degradation under a bounded budget,
and require a large storage space to meet state-of-the-art performance \cite{lama, vclimb}. Further, dense temporal sampling is necessary to maintain temporal association and long-term context in long videos \citep{htd1}; thus, heavy temporal down-sampling due to their frame selection strategies may explain the lower performance.

\subsection{Memory Budget Comparisons}
For the proposed method, we report the total memory buffer size and its equivalent size when storing raw pixel frames (table \ref{table:Tab2}). In table \ref{tab:method_mem_perf}, we show the total number of frames stored at the fixed memory budget per baseline. As described in section \ref{sec:incremental-formulation}, the proposed method only stores the autoencoder from the immediately-previous and the current tasks, thus using constant additional storage for models weights. We describe and report the average evaluation accuracy under various memory budgets (both higher and lower) for our method and baselines in figure \ref{fig:plot}. We note that we only show results on the better-performing baselines in this figure (for instance, \cite{lama} has a 2-4x lower memory budget and superior performance over \cite{vclimb}, reported in \cite{lama}). From this performance memory plot, we see that our method requires significantly less memory to achieve strong performance compared to prior art. Our method's memory budget is well within the capacity of modern hardware.
\subsection{Ablation Experiments}
\label{abl}
\begin{figure}[t]
\centering \scriptsize
\includegraphics[width=\columnwidth]{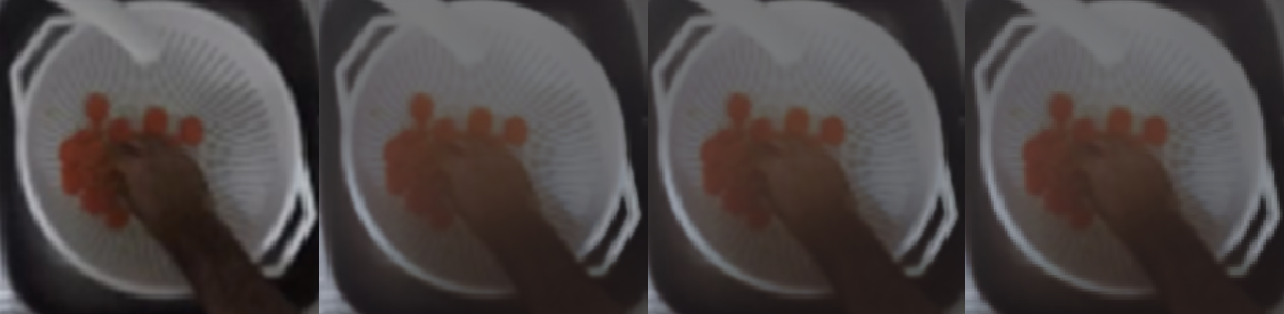}
    \caption{Original frame and our model's reconstructed frames.}
    \label{fig:wash}
\end{figure}

From tables \ref{table:tab00} and \ref{tab:video-length-ek100}, we observe that the method’s performance remains consistent with increasing video length up to several hundred hours (described in C.1, app.).
To understand the limits, we note that the method's performance will degrade due to forgetting, when the rehearsal buffer is unable to store data samples. To quantify the maximum video length beyond which our method’s performance degrades, we attempt to quantify an upper bound on the rehearsal buffer’s storage in Mb. In fig. \ref{fig:plot}, we see that \cite{lama} doesn't achieve stable performance under a limited memory budget, and in contrast, \cite{mem_cp} requires 20 Gb for comparable performance in K-700. If one assumes 20 Gb as an upper bound, our method can process 5470 hours of video length. We report results with a lower compression rate in table \ref{tab:tabxx}, and only see a small effect on our method's performance. In C.1 (app.), we report ablations with increased training per task and different class split per task. We show a close-up of a reconstructed frame in fig. \ref{fig:wash} (original, then reconstruction at stages 1-3), and observe that the details are not lost to any large extent. Further, in table \ref{tab:comparison}, we show the performance and memory tradeoffs (linear vs. constant) for the baselines in fig. \ref{fig:differences}, which largely follow the same trend.

\begin{table}[t]
\centering\scriptsize \setlength{\tabcolsep}{1pt}
\begin{tabular*}{\columnwidth}
{lcccccccccc}
\toprule
\multirow{2}{*}{\textbf{Method}} 
& \multicolumn{5}{c}{\textbf{Kinetics-700}} 
& \multicolumn{5}{c}{\textbf{Epic-Kitchens-100}} \\
\cmidrule(lr){2-6} \cmidrule(lr){7-11}
& Train $\uparrow$ & Eval $\uparrow$ & AvgF $\downarrow$ & Mem (Mb) & 
& Train $\uparrow$ & Eval $\uparrow$ & AvgF $\uparrow$ & Mem (Mb) & 
\\
\midrule
\textbf{CRAM} &  44.60 & 38.80 & 15.20 & 1404 &  &  32.60 &  28.10 & 21.60  & 1464 &  \\
\midrule
DF           & 45.1 & 37.5 & 14.6 & 13779 &  & 30.1 & 27.0  & 20.2  & 13089 &  \\ 
\midrule
CV    & 38.18 & 26.5  & 60.4 & 1029 &  & 26.6 & 19.1 & 62.1 & 1089 &  \\
\midrule
SGD             &  35.10 & 10.1 & 80.1 &  375 &  & 23.7 & 5.8 & 85.8 & 375 &  \\
\midrule
IID             &  70.1 & 58.2 & -  & - & - & 46.1 & 37.2 & - &  - & - \\
\bottomrule
\end{tabular*}
\caption{Ours (CRAM) vs. Fig. 2 baselines on K-700 and EK-100, with incremental settings Drift-Free, Compressed Vision, naive SGD (all from Sec. 5.3), and IID baseline.
We show Train/Eval acc., average forgetting (AvgF), and Memory (Mb).
}
\label{tab:comparison}
\end{table}

\section{Conclusion}
In this work we presented a method, named CRAM, to perform continual learning over long-videos, mitigating catastrophic forgetting.
Video CL poses considerable challenges, one of them being the high memory requirements. We propose to use compressed vision as a way to increase substantially the buffer size used for rehearsal in CL, and highlight the need to devise an appropriate strategy to deal with the representation drift of the compressor (i.e. codes become stale compared to the most recent compressor state).
We demonstrate encouraging results in 2 large-scale video datasets, Epic-Kitchens-100 and Kinetics-700.
We also study 2 different settings of CL, with pre-training and from scratch.
We believe that compressed vision can play an important role in scaling up methodologies developed for images and adapt them to videos.
In future work we would like to explore even more long-duration videos, and other tasks that go beyond action classification.

\paragraph{Acknowledgements.} The authors acknowledge the support of the Royal Society (RG\textbackslash R1\textbackslash 241385), Toyota Motor Europe (TME), and EPSRC (VisualAI, EP/T028572/1).

%
%
\bibliographystyle{ieeenat_fullname}
\bibliography{main}

\appendix
\section{Video Datasets Comparisons}
\label{sec:intro}

\begin{table*}[t]
\centering
\begin{tabular}{|p{2cm}|p{2cm}|p{2cm}|p{2cm}|p{2cm}|p{2.5cm}|}
\hline
\textbf{Dataset}           & \textbf{Longest Video Length (secs)} & \textbf{Average Video Length (secs)} & \textbf{\# of Object / Action Categories} & \textbf{Video understanding Setting} & \textbf{Used In} \\ 
\hline
Something-Something V2     & 6                                    & 4-6                                   & 174                        & short, fine-grained            & FrameMaker\citep{framemaker}, ST-Prompt \citep{st-prompt} \\ \hline 
HMDB51                     & 6                                    & 6                                     & 51                         & short                          &FrameMaker\cite{framemaker}, ST-Prompt\cite{st-prompt} \\ \hline
UCF101                     & 8                                    & 5-7                                   & 101                        & short                          &FrameMaker\cite{framemaker}, ST-Prompt\cite{st-prompt}, SMILE\cite{lama} \\ \hline
Kinetics (400/700)     & 20                                   & 10                                    & 400 / 700           & short                          & SMILE\cite{lama}, vCLIMB\cite{vclimb}, \bf{Ours} \\ \hline
ActivityNet                & 600 (10 mins)                       & 120                                   & 203                        & short                          & SMILE\cite{lama}, vCLIMB\cite{vclimb}, DPAT\cite{dpat} \\ \hline
Epic-Kitchens-100          & 5400 (1.5 hrs)                      & 900-1200 (15-20 mins)                 & 331                        & long, fine-grained             & DPAT\cite{dpat} (concurrent work), \bf{Ours} \\ \hline
\end{tabular}
\caption{Summary of video datasets: The following table describes each video dataset with the length of its longest video (column 2), average length (column 3), classification and temporal complexity in its video understanding setting (column 4, 5), and the respective CL works these datasets are used in (column 6).}
\label{table:tab22}
\end{table*}

\subsection{Dataset Details}
\paragraph{Epic-Kitchens-100} The average video length is 20 minutes, longest video length is 1.5 hours and shortest video length is 5 minutes. Total video footage length is 100 hours. Each video is at 25 frames per second.
We further describe the dataset annotations. Each video is associated with a participant and video identifier. Each video is split into a block of frames (segment) with a start and a stop timestamp, and indicated with the start and stop frame. A video segment is labeled with all the noun categories present in it (so multiple labels per clip). The labeling is at the video segment level. There are a total of 331 noun classes covering various nouns involved in kitchen actions (including everyday equipment). Smooth transitions between classes are ensured by presenting the segments to the models chronologically.

\begin{table*}[h!]
\centering
\begin{tabular}{|c|c|c|c|c|c|c|p{3cm}|}
\hline
\textbf{participant id} & \textbf{video id} & \textbf{start time} & \textbf{stop time} & \textbf{nouns} & \textbf{noun\_classes} \\ \hline
P01                      & P01\_01            & 00:29.22              &00:31.32                          & {[}``fridge''{]}         & {[}12{]}              \\ \hline
P01                      & P01\_01            &09:07.40              &09:09.01                           & {[}``container'', ``fridge''{]} & {[}21, 12{]}         \\ \hline
P02                      & P02\_108           &00:43.83              &00:45.92                         & {[}``biscuit'', ``cupboard''{]} & {[}104, 3{]}         \\ \hline
\end{tabular}
\caption{Example annotations from Epic-Kitchens-100 dataset}
\end{table*}

\paragraph{Kinetics-700} The average video length is 10 seconds, longest video length is 15 seconds and shortest video length is 7 seconds. Each video is at 25 frames per second.
There are 700 classes in total, and each class is also associated with an integer label (which is an integer value from 0 to 699). Each video is associated with a class label.

\begin{table*}[h!]
\centering
\begin{tabular}{|l|l|c|c|}
\hline
\textbf{label}               & \textbf{youtube\_id} & \textbf{start\_time} & \textbf{stop\_time} \\ \hline
``baking cookies''             & JJWwLganiil          & 31                   & 41                  \\ \hline
``gymnastics tumbling''        & 5KbfOS44-gM          & 49                   & 59                  \\ \hline
``writing''                    & iYcARQA6VIU          & 0                    & 10                  \\ \hline
``wrapping present''           & Qo5lspgmqPU          & 167                  & 177                 \\ \hline
\end{tabular}
\caption{Example annotations from Kinetics-700 dataset}
\end{table*}

\subsubsection{Dataset Diversity: Qualitative Analysis}
Epic Kitchens-100 (EK-100) and Kinetics-700 (K-700) cover a wide diversity in the videos both within and across the continual learning tasks.

\paragraph{Kinetics-700}
In K-700, video diversity comes from environmental context changes (eg: swimming / water, skiing / snow), range of motion and tools (e.g., paddleboarding vs. birdwatching), gestures (eg: teaching in a class vs poses during dancing), to name a few. In addition, for each action category in the dataset, the scene and protagonists vary.

\paragraph{Epic-Kitchens-100}
In EK-100, video diversity within each task comes from the same participant shooting at various day times in their kitchen, functionally repurposing various objects, variable scene length and shot type (based on the action performed), objects under multi-viewpoints, partial or full occlusion when captured temporally.
In EK-100, video diversity across tasks comes from new and culturally-diverse participants in their respective kitchens and cities. This leads to environmental, cinematography changes and intra-category variations for new or previously-seen objects and actions.

So, both short and long videos and varying amounts of action and scene changes are covered in each dataset. Further, we ensured that our model is presented with gradual complexity within and across the tasks ensuring smooth transitions (and by presenting data chronologically, wherever applicable) while preserving the diversity.

\begin{figure}    
    \includegraphics[width=0.5\textwidth]{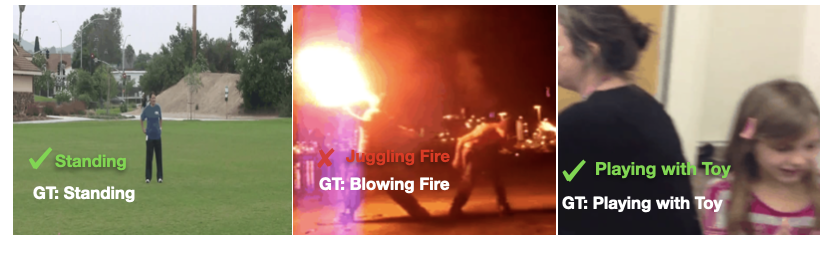}
    \caption{Visualization of the long-video continual learning setting on Kinetics-700 dataset with action classification.}
    \label{fig:kin}
\end{figure}
\section{Implementation Details}
\subsection{Average forgetting metric (AvgF)}
Let $a_{i,t}$  be accuracy on task i of the model that was trained on $t$ tasks, where $i < t$. Average forgetting measures how much performance has degraded across the first $t-1$ tasks. To do so, this metric uses the difference between  best-obtained performance of the desired task and the performance obtained from the current incremental learner.

\begin{alignat*}{1}
F_{t} & =\frac{1}{t-1}\sum_{1}^{t-1}f_{i,t}\\
 & \text{where}\quad f_{i,t}=\max_{q<t}\left(a_{i,q}-a_{i,t}\right) \quad \text{or}\quad f_{i,t}=a_{i,i}-a_{i,t}
\label{eq:eq11}
\end{alignat*}

\subsection{Baseline Details}

GDumb \citep{gd22} maintains a randomly-sampled RGB memory buffer. It stores all samples until the buffer is full and then stops storing.

REMIND \citep{mem_cp} proposes a compression technique using a two-stage process. In the first stage, it compresses the current input. This stage is analogous to the compression phase in our method. In the second-stage, it reconstructs a subset of previously compressed representations, and mixes them with the current input. It then updates the plastic weights of the network with this mixture. The second stage is analogous to decompression phase and rehearsal in our method to maintain stability of learned and new input representation.
We use their incremental batch version for the experimental results which is closest to our setting. A random selection strategy was used for REMIND. We directly apply their method by operating on RGB frames from videos instead of RGB samples from images. For base initialization phase, we use 20 classes for K-700 and 1 participant for EK-100 adapting their protocol as on ImageNet

SMILE \citep{lama} introduces a memory-based video CL baseline that maximizes the memory buffer usage by storing a single RGB frame per video. To combat the distribution shift between real video clips per CL task and in-memory images (represented as boring videos\citep{jc}), SMILE introduces a secondary loss. The method favors diversity of videos over temporal data per video. Their single-frame memory allows to directly apply image-based CL methods to the video domain. Similar to observations in GDumb \citep{gd22}, SMILE \citep{lama} also reports strong performance with a random sampling technique.

\section{Ablation Experiments} \label{abl1}

We perform ablation studies with increased training per task, a different split for classes per task and report more results from the ablation experiment described in section 5.8 in the main paper.

\subsection{Results}
In tables \ref{table:tab00} and \ref{tab:video-length-ek100}, we show how the performance of the proposed method varies with video length. In EK-100, in the pre-training setting, by the 5th task there is a total of $\sim$15 hours of video, and by the 10th task $\sim$25 hours. In the incremental-learning only setting, by the 20th task there are $\sim$50 hours, and by the 30th task, $\sim$80 hours. As we can see, the method’s performance remains consistent with increasing video length.

For the proposed method, we report experiment results with increased training epochs (40) per task in table \ref{tab:tab2}, (different than 30 epochs used in our main experiments), which shows a slight performance increase. This can be attributed to longer network training in the IID phase per task which allows for further loss reduction. 
We also perform experiments with a new split for classes per task in table \ref{tab:tab3}. For Kinetics-700, we try with 15 classes per task for 45 tasks in incremental setting, and 20 classes per task for 15 tasks in pre-training setting. For incremental setting, the training accuracy slightly increases due to fewer classes per task, however, the evaluation accuracy also reduces, indicating possible over-fitting. We see minimal effect in the pre-training setting, possibly due to stable class-agnostic representations learned during pre-training phase.
In tables \ref{table:tab01} and \ref{table:tab00}, we show how the performance of the proposed method varies with video length on Kinetics-700. As we can see, the method’s performance remains consistent with increasing video length.

\begin{table}[h!]  
\centering\setlength\tabcolsep{5pt}
\begin{tabular}{ccccccc}
\toprule
\multicolumn{3}{c}{Train.} & \multicolumn{3}{c}{Eval.}\tabularnewline
\midrule 
Task & 2 & 5 & 10 & 2 & 5 & 10 \tabularnewline
\midrule
Pretraining & 43.5 & 57.5 & 54.7 & 33.7 & 42.1 & 44.8 \tabularnewline
\bottomrule
\end{tabular}
\caption{Average training and evaluation accuracy with increasing video length, reported at the end of task $t$, on Kinetics-700 in the pretraining setting (Sec 5.2).}
\label{table:tab01}
\end{table}

\begin{table}[t]  
\centering\setlength\tabcolsep{5pt}
\begin{tabular}{ccccccc}
\toprule
\multicolumn{3}{c}{Train.} & \multicolumn{3}{c}{Eval.} \tabularnewline
\midrule 
Task & 10 & 20 & 30 & 10 & 20 & 30 \tabularnewline
\midrule 
Incremental & 37.5 & 41.8 & 40.3 & 25.8 & 33.1 & 35.7 \tabularnewline
\bottomrule
\end{tabular}
\caption{Average training and evaluation accuracy with increasing video length, reported at the end of task $t$, on Kinetics-700 in the incremental settings (Sec 5.2).}
\label{table:tab00}
\end{table}


\begin{table}[t]
\centering\setlength\tabcolsep{8pt}
\begin{tabular}{ccccc}
\toprule
 & \multicolumn{2}{c}{Kinetics-700} & \multicolumn{2}{c}{EpicKitchens-100}\\
 \cmidrule{2-3} \cmidrule(l){4-5}
 & Train. & Eval. & Train. & Eval.\\
\midrule 
Pretraining & 56.8 & 47.9 & 41.4 & 35.5\\
\midrule 
Incremental & 47.2 & 41.9 & 34.7 & 31.1 \\
\bottomrule
\end{tabular}
\caption{Our method's ablation with a different number of training epochs (40). Training (Train) and evaluation (Eval) performance reported above}
\label{tab:tab2}
\end{table}

\begin{table}[t]
\centering\setlength\tabcolsep{8pt}
\begin{tabular}{ccc}
\toprule
Setting & Train. & Eval. \\
\midrule 
{Pre-training} & 56.8 & 47.0
\\
\midrule
{Incremental} & 46.6 & 36.1
\\
\bottomrule
\end{tabular}
\caption{Our method's ablation with a different split as explained in \eqref{abl1}. Training (Train) and evaluation (Eval) accuracy on Kinetics-700}
\label{tab:tab3}
\end{table}

\begin{table}[t]
\centering\footnotesize
\setlength\tabcolsep{5pt}
\begin{tabular}{ccccccc}
\toprule 
Setting & \multicolumn{3}{c}{Pretraining} & \multicolumn{3}{c}{Incremental}\tabularnewline
\cmidrule{2-4} \cmidrule(l){5-7}
Task & 2 & 5 & 10 & 10 & 20 & 30\tabularnewline
\midrule
\multirow{1}{*}{Training} & 36.9 & 34.1 & 38.9 & 28.5 & 31.2 & 29.7\tabularnewline
\midrule 
\multirow{1}{*}{Evaluation} & 31.2 & 29.8 & 34.8 & 27.5 & 24.6 & 32.3\tabularnewline
\bottomrule
\end{tabular}
\caption{Average training and evaluation accuracy with increasing video length, reported at the end of task $t$, on Epic-Kitchens-100 in pretraining and incremental settings (Sec 5.2)}
\label{tab:video-length-ek100}
\end{table}

\begin{table}[t]
\centering\footnotesize
\setlength\tabcolsep{4pt}
\begin{tabular}{ccccc}
\toprule
Compression & \multicolumn{2}{c}{Kinetics-700} & \multicolumn{2}{c}{EpicKitchens-100}\\
 \cmidrule{2-3} \cmidrule(l){4-5}
Setting & Train. & Eval. & Train. & Eval.\\
\midrule 
Pretraining & 56.9 & 47.4 & 41.0 & 34.5\\
\midrule 
Incremental & 46.0 & 40.1 & 33.6 & 29.0 \\
\bottomrule
\end{tabular}
\caption{Average training (Train) and evaluation (Eval) accuracy for our method with a different compression rate (50$\times$).}
\label{tab:tabxx}
\end{table}

\end{document}